\documentclass[letterpaper]{article}
\usepackage{ijcai18}
\usepackage{multirow}
\usepackage{dblfloatfix}
\usepackage{times}
\usepackage{graphicx}
\usepackage{amsmath}
\usepackage{amsfonts}
\usepackage{color}
\usepackage{tabularx,threeparttable}
\usepackage{multirow}
\usepackage{booktabs}
\usepackage{comment}
\usepackage[pass]{geometry}
\usepackage[small]{caption}

\usepackage{fancyhdr}

\title{Unsupervised Cross-Modality Domain Adaptation of ConvNets for \\ Biomedical Image Segmentations with Adversarial Loss}
\author{Qi Dou$^{1}$\thanks{Authors contributed equally.}, Cheng Ouyang$^{2}$$^\ast$, Cheng Chen$^{1}$, Hao Chen$^{1}$$^{,}$$^{3}$ {\normalfont and} Pheng-Ann Heng$^{1}$ \\
 $^{1}$ { Department of Computer Science and Engineering, The Chinese University of Hong Kong } \\
 $^{2}$ { Department of Electrical Engineering and Computer Science, University of Michigan} \\
 $^{3}$ { Imsight Medical Technology Inc., Shenzhen, China}\\
{qdou@cse.cuhk.edu.hk, couy@umich.edu, \{cchen,hchen,pheng\}@cse.cuhk.edu.hk}}

\begin{document}

\maketitle

\begin{abstract}
\label{abs}
Convolutional networks (ConvNets) have achieved great successes in various challenging vision tasks.
However, the performance of ConvNets would degrade when encountering the domain shift.
The domain adaptation is more significant while challenging in the field of biomedical image analysis, where cross-modality data have largely different distributions.
Given that annotating the medical data is especially expensive, the supervised transfer learning approaches are not quite optimal.
In this paper, we propose an unsupervised domain adaptation framework with adversarial learning for cross-modality biomedical image segmentations.
Specifically, our model is based on a dilated fully convolutional network for pixel-wise prediction. 
Moreover, we build a plug-and-play domain adaptation module (DAM) to map the target input to features which are aligned with source domain feature space.
A domain critic module (DCM) is set up for discriminating the feature space of both domains.
We optimize the DAM and DCM via an adversarial loss without using any target domain label.
Our proposed method is validated by adapting a ConvNet trained with MRI images to unpaired CT data for cardiac structures segmentations, and achieved very promising results.

\end{abstract}

\section{Introduction}
\label{sec:intro}

Deep convolutional networks (ConvNets) have demonstrated great achievements in recent years, achieving state-of-the-art or even human-level performance on various computer vision challenging problems, such as image recognition, semantic segmentation as well as biomedical image diagnosis~\cite{he2016deep,esteva2017dermatologist}.
Typically, the deep networks are trained and tested on datasets where all the samples are drawn from the same probability distribution.
However, it has been observed that established models would under-perform when tested on samples from a related but not identical new target domain~\cite{shimodaira2000improving}.

\begin{figure}
\centering
\includegraphics[width=0.48\textwidth]{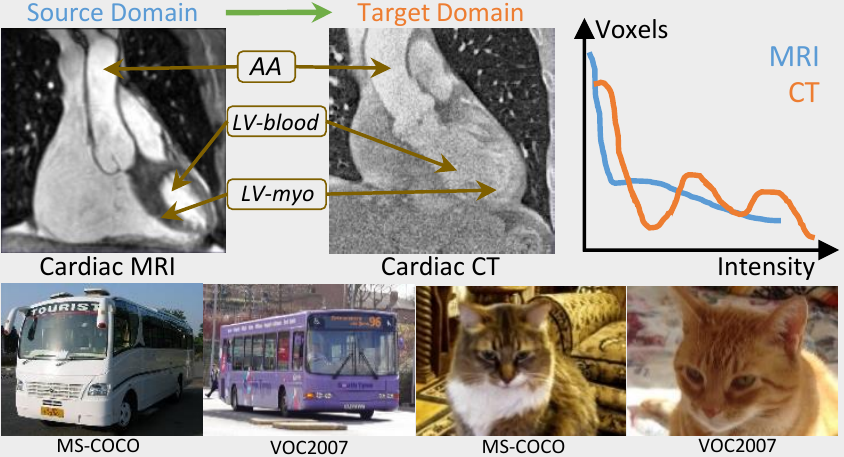}
\vspace{-5mm}
\caption{Illustration of severe domain shift existing in cross-modality biomedical images. The appearances of the anatomical structures (AA: ascending aorta, LV-blood: left ventricle blood cavity, LV-myo: left ventricle myocardium) would vary significantly on MRI and CT images. Compared with natural image datasets (see bottom examples), domain adaptation for cross-modality medical images encounter more challenges.}
\vspace{-4mm}
\label{fig:domain_shift}
\end{figure}

The existence of \textit{domain shift} is common in real-life applications~\cite{gretton2009covariate,torralba2011unbiased}.
The semantic class labels are usually shared between domains, whereas the distributions of data are different.
In the field of biomedical image analysis, this issue is even more obvious.
Unlike natural images which are generally taken by optical devices, medical radiological images are acquired by different imaging modalities such as Computed Tomography (CT) and Magnetic Resonance Imaging (MRI).
Data distributions of these modalities mismatch significantly, due to their different principles of imaging physics.
The appearance of anatomical structures are distinct across radiology modalities, with obviously different intensity histograms.
In Fig.~\ref{fig:domain_shift}, we illustrate the severe domain shift between MRI/CT data.
In comparison with examples from natural datasets, domain adaptation for cross-modality medical data is more challenging.

To tackle this issue, domain adaptation methods have been studied to generalize the learned models~\cite{patel2015visual}.
The domain of labeled training data is termed as~\textit{source domain}, and the test dataset is called~\textit{target domain}.
A straight-forward solution is transfer learning, i.e., fine-tuning the models learned on source domain with extra labeled data from the target domain~\cite{pan2010survey}. However, the annotation is prohibitively time-consuming and expensive, especially for those biomedical datasets.
Alternatively, the unsupervised domain adaptation methods are more feasible, given that these scenarios transfer knowledge across domains without using additional target domain labels.
Advanced studies in this direction have taken advantage of adversarial training to implicitly learn the feature mapping between domains, and achieved remarkable success in natural datasets~\cite{ganin2016domain,tzeng2017adversarial}.

Currently, for biomedical images, how to effectively generalize ConvNets across domains has not yet been fully studied.
A representative work is~\cite{kamnitsas2017unsupervised} which conducted unsupervised domain adaptation for brain lesion segmentation and achieved promising results.
However, their source and target domains are relatively close, given that both are MRI datasets although acquired with different scanners.
Adapting ConvNets between cross-modality radiology images with a huge domain shift is more compelling for clinical practice, but has not been explored yet.

In this paper, we propose a novel cross-modality domain adaptation framework for medical image segmentations with unsupervised adversarial learning.
To transfer the established ConvNet from source domain (MRI) to target domain (CT) images, we design a plug-and-play domain adaptation module (DAM) which implicitly maps the target input data to the feature space of source domain.
Furthermore, we construct a discriminator which is also a ConvNet termed as domain critic module (DCM) to differentiate the feature distributions of two domains. 
Adversarial loss is derived to train the entire domain adaptation framework in an unsupervised manner, by placing the DAM and DCM into a minimax two-player game.
Our main contributions are:

\begin{itemize}
\item We pioneer cross-modality domain adaptation for medical image segmentation using deep ConvNets.
A flexible plug-and-play framework is designed to transfer a MRI segmenter to CT data via feature-level mapping.
\item We optimize our framework with unpaired MRI/CT images via adversarial learning in an unsupervised manner, eliminating the cost of labeling extra medical datasets.
\item Extensive experiments with promising results on cardiac segmentation application have validated the feasibility of radiology cross-modality domain adaptation, as well as the effectiveness of our approach towards this task.
\end{itemize}

\section{Related Work}

Domain adaptation aims to confront the performance degradation caused by any distribution change occurred after learning a classifier.
For deep learning models, this situation also applies, and a trend of studies have been conducted to map the target input to the original source domain or its feature space.
In this section, we first present related works of unsupervised domain adaptation that achieved promising results on natural image datasets.
Next, we review the recent studies on domain adaptation for medical image segmentations using ConvNets.

Most prior studies on unsupervised domain adaptation focused on aligning the distributions between domains in feature space, by minimizing measures of distance between features extracted from the source and target domains.
For example, the Maximum Mean Discrepancy (MMD) was minimized together with a task-specific loss to learn the domain-invariant and semantic-meaningful features in~\cite{tzeng2014deep}.
The correlations of layer activations between the domains were aligned in the study of~\cite{sun2016deep}.
Based on this,~\cite{wang2017deep} further extended the work and minimized domain difference based on both the first and second order information between source and target domains.
Alternatively, with the emergence of generative adversarial network (GAN) \cite{goodfellow2014generative} and its powerful extensions \cite{radford2015unsupervised,arjovsky2017wasserstein}, the mapping between domains were implicitly learned via the adversarial loss. 
The~\cite{ganin2016domain} proposed to extract domain-invariant features by sharing weights between two ConvNet classifiers.
Later, the~\cite{tzeng2017adversarial} introduced a more flexible adversarial learning method with untied weight sharing, which helps effective learning in the presence of larger domain shifts.
Another GAN based direction of solution is to learn a transformation in the pixel space~\cite{Bousmalis_2017_CVPR}, adapting the source-domain images to appear as if drawn from the target domain.

In the field of medical image analysis using deep learning, domain adaptation is also an important topic to generalize learned models across data acquired from different imaging protocols.
Transfer learning with network fine-tuning strategies has been experimentally studied by~\cite{ghafoorian2017transfer} on the brain lesion segmentation application.
Although the amount was small, annotations from target domain were still required in their scenario.
The latest study on medical data that is closely related to our work is~\cite{kamnitsas2017unsupervised}, which performed unsupervised domain adaptation for brain lesion segmentation.
Their ConvNets learned domain-invariant features on images, with an adversarial loss serving as the supervision for feature extraction. The results were inspiring and demonstrated the efficacy of adversarial loss for unsupervised domain adaptation on medical datasets.
However, their source and target domains are relatively close, because both were MRI datasets.
Although acquired with different scanners and imaging protocols, the images were from the same modality and the domain shift was not dramatic.
In contrast, our problem setting, i.e., adapting a ConvNet trained on MRI data to CT images, is novel but more adventurous and challenging, since our domain shift is more severe.

\section{Methods}

\begin{figure*}[t]
\label{fig:overview}
\centering
\includegraphics[width=\textwidth]{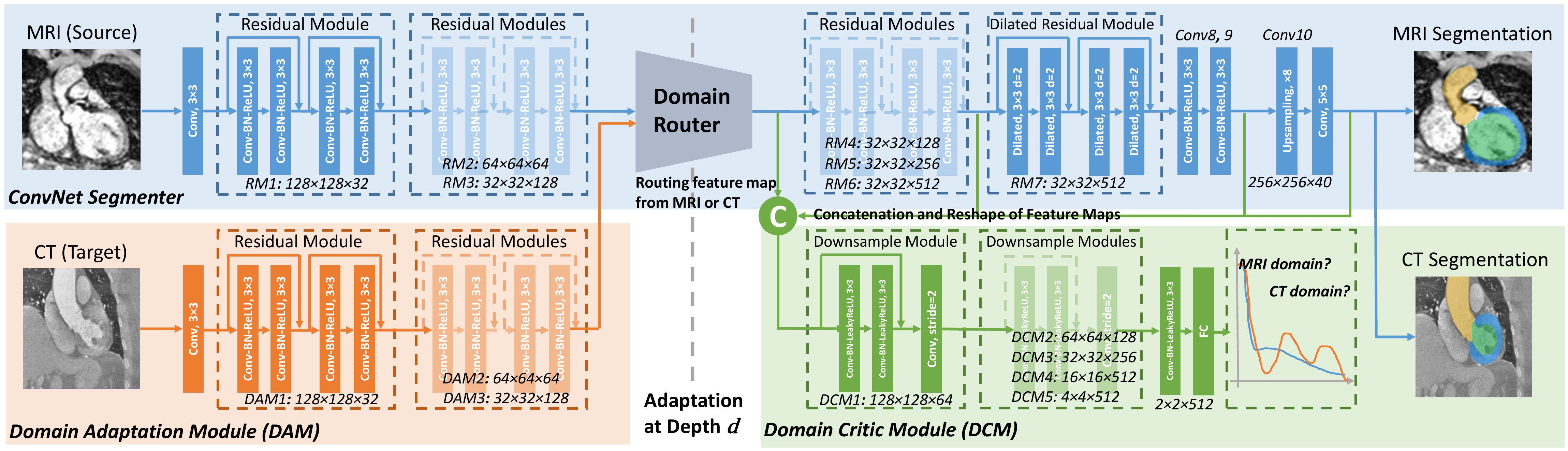}
\vspace{-6mm}
\caption{Overview of our proposed plug-and-play framework for cross-modality domain adaptation. The DAM and DCM are optimized via adversarial learning. During inference, the domain router is used for routing feature maps of different domains.}
\vspace{-4mm}
\end{figure*}

The Fig.~2 presents our proposed framework for unsupervised cross-modality domain adaptation in biomedical image segmentation. 
Based on a standard ConvNet segmenter, we construct a plug-and-play domain adaptation module (DAM) and a domain critic module (DCM) to form adversarial learning. Details of network architecture, adaptation method, adversarial loss and training strategies are elaborated in this section.

\subsection{ConvNet Segmenter Architecture}

With the labeled dataset of $N^s$ samples from source domain, denoted by $X^s \! = \! \{(x_1^s,y_1^s),...,(x_{N^s}^s,y_{N^s}^s)\}$, we conduct supervised learning to establish a mapping from the input image to the label space $Y^s$.
In our setting, the $x_i^s$ represents the sample (pixel or patch) of medical images and $y_i^s$ is the category of anatomical structures.
For the ease of denotation, we omit the index $i$ in the following, and directly use $x^s$ and $y^s$ to represent the samples and labels from the source domain.

The mapping $M^s$ from input to the label space is implicitly learned in the form of a segmentation ConvNet.
The backbone of our segmenter is the residual network for pixel-wise prediction of biomedical images.
We employ the dilated residual blocks~\cite{yu2017dilated} to extract representative features from a large receptive field while preserving the spatial acuity of feature maps.
More specifically, the image is firstly input to a Conv layer, then forwarded to 3 residual modules (termed as RM, each consisting of 2 stacked residual blocks) and downsampled by a factor of 8.
Next, another three RMs and one dilated RM are stacked to form a deep network.
To enlarge receptive field for extracting global semantic features, 4 dilated convolutional layers are used in RM7 with a dilation factor of 2.
For dense predictions in our segmentation task, we conduct upsamling at layer \textit{Conv10},
which is followed by $5 \! \times \! 5$ convolutions to smooth out the feature maps.
Finally, a softmax layer is used for probability predictions of the pixels.

The segmentation ConvNet using labeled data from source domain is optimized by minimizing the hybrid loss $\mathcal{L}_\text{seg}$ composed of the multi-class cross-entropy loss and the Dice coefficient loss~\cite{milletari2016v}.
Formally, we denote $y_{i,c}^s$ for binary label regarding class $c \! \in \! C$ in sample $x_i^s$, its probability prediction is $\hat{p}_{i,c}^s$, and the label prediction is $\hat{y}_{i,c}^s$, the source domain segmenter loss function is as follows:
\vspace{-2mm}
\begin{equation}
\small
\begin{split}
\mathcal{L}_{\text{seg}} % (Y^s, \hat{Y}^s)
	= & -\sum \limits_{i=1}^{N^s} \sum \limits_{c \in C} w^s_{c} \cdot y_{i,c}^s \log(\hat{p}_{i,c}^s) \\
	  & -\lambda \sum \limits_{c \in C} \frac{ \sum_{i=1}^{N^s} 2 y_{i,c}^s \hat{y}_{i,c}^s }{  \sum_{i=1}^{N^s} y_{i,c}^s y_{i,c}^s + \sum_{i=1}^{N^s} \hat{y}_{i,c}^s \hat{y}_{i,c}^s },
\end{split}
\end{equation}
where the first term is the cross-entropy loss for pixel-wise classification, with $w_{c}^s$ being a weighting factor to cope with the issue of class imbalance. The second term is the Dice loss for multiple cardiac structures, which is commonly employed in biomedical image segmentation problems.
We combine the two complementary loss functions to tackle the challenging heart segmentation task.
In practice, we also tried to use only one type of loss, but the performance was not quite high.

\subsection{Plug-and-Play Domain Adaptation Module}

When the ConvNet is learned on the source domain, our goal is to generalize it to a target domain.
In transfer learning, the last several layers of the network are usually fine-tuned for a new task with new label space.
The supporting assumption is that early layers in the network extract low-level features (such as edge filters and color blobs) which are common for vision tasks.
Those upper layers are more task-specific and learn high-level features for the classifier~\cite{zeiler2014visualizing,yosinski2014transferable}.
In this case, labeled data from target domain are required to supervise the learning process.
Differently, we use unlabeled data from the target domain, given that labeling dataset is time-consuming and expensive.
This is critical in clinical practice where radiologists are willing to perform image computing on cross-modality data with as less extra annotation cost as possible.
Hence, we propose to adapt the ConvNet with unsupervised learning.

In our segmenter, the source domain mapping $M^s$ is layer-wise feature extractors composing stacked transformations of $\{M^s_{l_1},...,M^s_{l_n}\}$, with the $l$ denoting the network layer index. Formally, the predictions of labels are obtained by:
\begin{equation}
\small
\hat{y}^s = M^s(x^s)= M^s_{l_1:l_n}(x^s) = M^s_{l_n}  \circ ... \circ M^s_{l_1} (x^s).
\end{equation}

For domain adaptation, the label space of source and target domains are identical, i.e., we segment the same anatomical structures from medical MRI/CT data.
Our hypothesis is that the distribution changes between the cross-modality domains are primarily low-level characteristics (e.g., gray-scale values) rather than high-level (e.g., geometric structures).
The higher layers (such as $M^s_{l_n}$) are closely in correlation with the class labels which can be shared across different domains.
In this regard, we propose to reuse the feature extractors learned in higher layers of the ConvNet, whereas the earlier layers are updated to conduct distribution mappings in feature space for our unsupervised domain adaptation.

For the input from target domain $x^t$, we propose a domain adaptation module denoted by $\mathcal{M}$ that maps $x^t$ to the feature space of the source domain.
We denote the adaptation depth by $d$, i.e., the layers earlier than and including $l_d$ are replaced by DAM when processing the target domain images.
In the meanwhile, the source model's upper layers are frozen during domain adaptation learning and reused for target inference.
Formally, the predictions for target domain is as:
\vspace{-1mm}
\begin{equation}
\small
\begin{aligned}
\hat{y}^t = M^s_{l_{d+1}:l_n} \circ \mathcal{M}(x^t)= M^s_{l_n}  \circ ... \circ M^s_{l_{d+1}} \circ \mathcal{M} (x^t),
\end{aligned}
\end{equation}
where $\mathcal{M}(x^t) = \mathcal{M}_{l_1:l_d}(x^t)  = \mathcal{M}_{l_d}  \circ ... \circ \mathcal{M}_{l_1} (x^t)$ represents the DAM which is also a stacked ConvNet.
Overall, we form a flexible plug-and-play domain adaptation framework.
During the test inference, the DAM directly replaces the early $d$ layers of the model trained on source domain.
The images of target domain are processed and mapped to deep learning feature space of source domain via the DAM.
These adapted features are robust to the cross-modality domain shift, and can be mapped to the label space using those high-level layers established on source domain.
In practice, the ConvNet configuration of the DAM is identical to $\{M^s_{l_1},...,M^s_{l_d}\}$.
We initialize the DAM with trained source domain model and fine-tune the parameters in an unsupervised manner with adversarial loss. 

\subsection{Learning with Adversarial Loss}

We propose to train our domain adaptation framework with adversarial loss via unsupervised learning. 
The spirit of adversarial training roots in GAN, where a generator model and a discriminator model form a minimax two-player game.
The generator learns to capture the real data distribution; and the discriminator estimates the probability that a sample comes from the real training data rather than the generated data.
These two models are alternatively optimized and compete with each other, until the generator can produce real-like samples that the discriminator fails to differentiate. 
For our problem, we train the DAM, aiming that the ConvNet can generate source-like feature maps from target input.
Hence, the ConvNet is equivalent to a generator from GAN's perspective. 

Considering that accurate segmentations come from high-level semantic features, which in turn rely on fine-patterns extracted by early layers, we propose to align multiple levels of feature maps between source and target domains (see Fig.~2).
In practice, we select several layers from the frozen higher layers, and refer their corresponding feature maps as the set of $F_H (\cdot)$ where $H \! = \! \{k, ... ,q\}$ being the set of selected layer indices.
Similarly, we denote the selected feature maps of DAM by $\mathcal{M}_A( \cdot )$ with the $A$ being the selected layer set.
In this way, the feature space of target domain is $(\mathcal{M}_A(x^t), F_H(x^t))$ and the $(M^s_A(x^s), F_H(x^s))$ is their counterpart for source domain.
Given the distribution of $(\mathcal{M}_A(x^t), F_H(x^t)) \! \sim \! \mathbb{P}_g$, and that of $(M^s_A(x^s), F_H(x^s)) \! \sim \! \mathbb{P}_s$, the distance between these two domain distributions which needs to be minimized is represented as $W(\mathbb{P}_s,\mathbb{P}_g)$.
For stabilized training, we employ the Wassertein distance~\cite{arjovsky2017wasserstein} between the two distributions as follows:
\begin{equation}
\small
\vspace{-1mm}
W(\mathbb{P}_s, \mathbb{P}_g)=\inf \limits_{\gamma \sim \prod(\mathbb{P}_s, \mathbb{P}_g)} \mathbb{E}_{(\text{x},\text{y})\sim\gamma}[\Vert \text{x} - \text{y} \Vert],
\label{equ:emdis}
%\vspace{-1mm}
\end{equation}
where $\prod(\mathbb{P}_s, \mathbb{P}_g)$ represents the set of all joint distributions $\gamma(\text{x},\text{y})$ whose marginals are respectively $\mathbb{P}_s$ and $\mathbb{P}_g$.

In adversarial learning, the DAM is pitted against an adversary: a discriminative model that implicitly estimates the $W(\mathbb{P}_s,\mathbb{P}_g)$.
We refer our discriminator as domain critic module and denote it by $\mathcal{D}$.
Specifically, our constructed DCM consists of several stacked residual blocks, as illustrated in Fig.~2.
In each block, the number of feature maps is doubled until it reaches 512, while their sizes are decreased.
We concatenate the multiple levels of feature maps as input to the DCM.
This discriminator would differentiate the complicated feature space between the source and target domains.
In this way, our domain adaptation approach not only removes source-specific patterns in the beginning but also disallows their recovery at higher layers~\cite{kamnitsas2017unsupervised}.
In unsupervised learning, we jointly optimize the generator $\mathcal{M}$ (DAM) and the discriminator $\mathcal{D}$ (DCM) via adversarial loss. 
Specifically, with $X^t$ being target set, the loss for learning the DAM is:
\begin{equation}
\vspace{-1mm}
\small
\begin{split}
\min \limits_{\mathcal{M}}  \mathcal{L}_{\mathcal{M}}&(X^t,\mathcal{D}) \! = \! \\ -\mathbb{E}&_{ (\mathcal{M}_{A}(x^t), F_{H}(x^t)) \sim \mathbb{P}_g} [\mathcal{D}(\mathcal{M}_{A}(x^t), F_{H}(x^t))].
\label{equ:Gloss}
\end{split}
\end{equation}
Furthermore, with the $X^s$ representing the set of source images, the DCM is optimized via:
\begin{equation}
\small
\begin{split}
&\min \limits_{\mathcal{D}} \mathcal{L}_{\mathcal{D}}(X^s, X^t, \mathcal{M}) = \\ 
&\mathbb{E}_{(\mathcal{M}_{A}(x^t), F_{H}(x^t))  \sim \mathbb{P}_g}[\mathcal{D}(\mathcal{M}_{A}(x^t), F_{H}(x^t))] \ - \\
&\mathbb{E}_{ (M^s_{A}(x^s), F_{H}(x^s))\sim \mathbb{P}_s}[\mathcal{D}(M^s_{A}(x^s),F_{H}(x^s))],  s.t. \  \Vert \mathcal{D} \Vert_{L\leq K},
\end{split}
\label{equ:Dloss}
\end{equation}
where $K$ is a constant that applies Lipschitz contraint to $\mathcal{D}$.

During the alternative updating of $\mathcal{M}$ and $\mathcal{D}$, the DCM outputs a more precise estimation of $W(\mathbb{P}_s, \mathbb{P}_g)$ between distributions of the feature space from both domains.
The updated DAM is more effective to generate source-like feature maps for conducting cross-modality domain adaptation.

\subsection{Training Strategies}
\label{subsec:training}

In our setting, the source domain is biomedical cardiac MRI images and the target domain is CT data.
All the volumetric MRI and CT images were re-sampled to the voxel spacing of $1 \! \times \! 1 \! \times \! 1$ mm$^3$ and cropped into the size of $256 \! \times \! 256 \! \times \! 256$ centering at the heart region. In preprocessing, we conducted intensity standardization for each domain, respectively.
Augmentations of rotation, zooming and affine transformations were employed to combat over-fitting.  
To leverage the spatial information existing in volumetric data, we sampled consecutive three slices along the coronal plane and input them to three channels.
The label of the intermediate slice is utilized as the ground truth when training the 2D networks.

We first trained the segmenter on the source domain data in supervised manner with stochastic gradient descent.
The Adam optimizer was employed with parameters as batch size of 5, learning rate of $1 \! \times \! 10^{-3}$ and a stepped decay rate of 0.95 every 1500 iterations.
After that, we alternatively optimized the DAM and DCM with the adversarial loss for unsupervised domain adaptation. 
Following the heuristic rules of training WGAN~\cite{arjovsky2017wasserstein}, we updated the DAM every 20 times when updating the DCM.
In adversarial learning, we utilized the RMSProp optimizer with a learning rate of $3\times 10^{-4}$ and a stepped decay rate of 0.98 every 100 joint updates, with weight clipping for the discriminator being 0.03.

\section{Experiment}
\subsection{Dataset and Evaluation Metrics}
\label{subsec:mat}

\begin{figure*}[!htp]
\label{fig:cmp}
\centering
\includegraphics[width=0.96\textwidth]{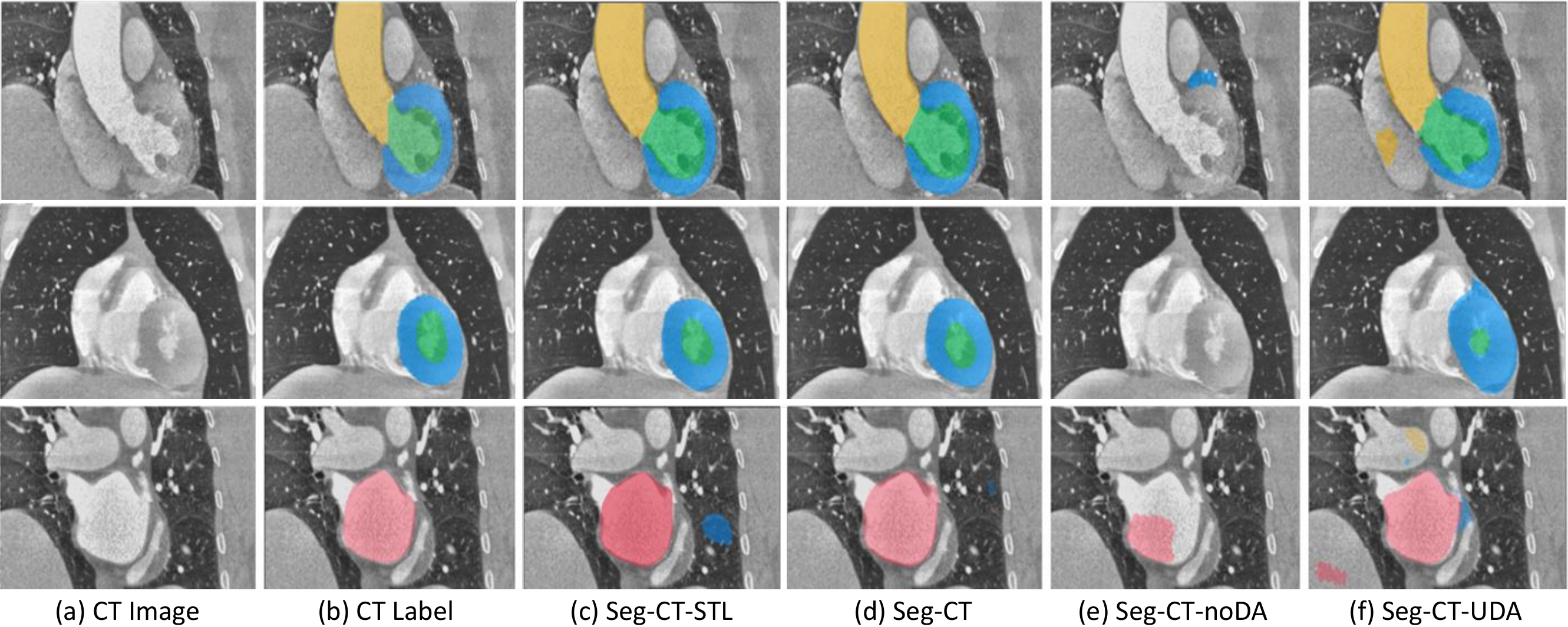}
\vspace{-3mm}
\caption{Results of different methods for CT image segmentations. Each row presents one typical example, from left to right: (a) raw CT slices (b) ground truth labels (c) supervised transfer learning (d) ConvNets trained from scratch (e) directly applying MRI segmenter on CT data (f) our unsupervised cross-modality domain adaptation results. The structures of AA, LA-blood, LV-blood and LV-myo are indicated by yellow, red, green and blue colors, respectively (best viewed in color).}
\vspace{-3mm}
\end{figure*}

We validated our proposed unsupervised cross-modality domain adaptation method for biomedical image segmentations on the public dataset of \textit{MICCAI 2017 Multi-Modality Whole Heart Segmentation}~\cite{zhuang2016multi}.
This dataset consists of unpaired 20 MRI and 20 CT images from 40 patients.
The MRI and CT data were acquired in different clinical centers.
The cardiac structures of the images were manually annotated by radiologists for both MRI and CT images.
Our ConvNet segmenter aimed to automatically segment four cardiac structures including the ascending aorta (AA), the left atrium blood cavity (LA-blood), the left ventricle blood cavity (LV-blood), and the myocardium of the left ventricle (LV-myo). 
For each modality, we randomly split the dataset into training (16 subjects) and testing (4 subjects) sets, which were fixed throughout all experiments.

For evaluation metrics, we followed the common practice to quantitatively evaluate the segmentation performance for automatic methods~\cite{dou20173d}.
The DICE coefficient $\!([\%])\!$ was employed to assess the agreement between the predicted segmentation and ground truth for cardiac structures.
We also calculated the average surface distance (ASD$[\text{voxel}]$) to measure the segmentation performance from the perspective of the boundary.
A higher Dice and lower ASD indicate better segmentation performance.
Both metrics are presented in the format of \emph{mean$\pm$std}, which shows the average performance as well as the cross-subject variations of the results.

\begin{table*}[t]
\label{tab:results}
\vspace{-3mm}
\centering
\small
\renewcommand{\arraystretch}{1.2}
\begin{center}
\begin{tabular}{ |m{3.46cm}|m{1.3cm}|m{1.3cm}|m{1.3cm}|m{1.3cm}|m{1.3cm}|m{1.3cm}|m{1.3cm}|m{1.3cm}|  }
\hline
Methods & \multicolumn{2}{c|}{AA} & \multicolumn{2}{c|}{LA-blood} & \multicolumn{2}{c|}{LV-blood} & \multicolumn{2}{c|}{LV-myo} \\
\cline{2-9}
&~~~ Dice & ~~~ ASD & ~~~ Dice & ~~~ ASD & ~~~ Dice & ~~~ ASD & ~~~ Dice  & ~~~ ASD \\
\hline

DL-MR~\cite{payermulti}  & 76.6$\pm$13.8 & ~~~~~~~  - & 81.1$\pm$13.8 & ~~~~~~~  -  & 87.7$\pm$7.7 & ~~~~~~~  -  & 75.2$\pm$12.1 & ~~~~~~~  - \\

DL-CT~\cite{payermulti}      & 91.1$\pm$18.4 &~~~~~~~   - & 92.4$\pm$3.6 & ~~~~~~~  -  & 92.4$\pm$3.3 &~~~~~~~   -  & 87.2$\pm$3.9 & ~~~~~~~  - \\

\hline
Seg-MRI    & 75.9$\pm$5.5  & 12.9$\pm$8.4  & 78.8$\pm$6.8 & 16.0$\pm$8.1   & 90.3$\pm$1.3 & ~~2.0$\pm$0.2   & 75.5$\pm$3.6 & ~~2.6$\pm$1.4 \\

Seg-CT     & 81.3$\pm$24.4 & ~~2.1$\pm$1.1  & 89.1$\pm$3.0 & 10.6$\pm$6.9   & 88.8$\pm$3.7 & 21.3$\pm$8.8  & 73.3$\pm$5.9 & 42.8$\pm$16.4\\

Seg-CT-STL & 78.3$\pm$2.8  & ~~2.9$\pm$2.0  & 89.7$\pm$3.6 & ~~7.6$\pm$6.7  & 91.6$\pm$2.2 & ~~4.9$\pm$3.2   & 85.2$\pm$3.3 & ~~5.9$\pm$3.8 \\

\hline
Seg-CT-noDA& 19.7$\pm$2.0  & 31.2$\pm$17.5 & 25.7$\pm$17.2& ~~8.7$\pm$3.3  & ~~0.8$\pm$1.3  & ~~~~~N/A  & 11.1$\pm$14.4& 31.0$\pm$37.6\\

\hline
Seg-CT-UDA~($d$=13) & 63.9$\pm$15.4 & \textbf{13.9$\pm$5.6} & 54.7$\pm$13.2 & 16.6$\pm$6.8 & 35.1$\pm$26.1 &\textbf{18.4$\pm$5.1} & 35.4$\pm$18.4 &\textbf{14.2$\pm$5.3}\\

Seg-CT-UDA~($d$=21) & \textbf{74.8$\pm$6.2}  & 27.5$\pm$7.6  & 51.1$\pm$11.2& 20.1$\pm$4.5   & \textbf{57.2$\pm$12.4}& 29.5$\pm$11.7 & \textbf{47.8$\pm$5.8} & 31.2$\pm$10.1\\

Seg-CT-UDA~($d$=31) & 71.9$\pm$0.5  & 25.8$\pm$12.5 & \textbf{55.2$\pm$22.9} & \textbf{15.2$\pm$8.2} & 39.2$\pm$21.8  & 21.2$\pm$3.9 & 34.3$\pm$19.1 & 24.7$\pm$10.5\\
\hline
\end{tabular}
\caption{Quantitative comparison of segmentation performance on cardiac structures between different methods. (Note: the - means that the results were not reported by that method.)}
\vspace{-5mm}
\end{center}
\end{table*}

\subsection{Experimental Settings}
In our experiments, the source domain is the MRI images and the target domain is the CT dataset.
We demonstrated the effectiveness of the proposed unsupervised cross-modality domain adaptation method with extensive experiments.
We designed several experiment settings: 
1) training and testing the ConvNet segmenter on source domain (referred as \textit{Seg-MRI});
2) training the segmenter from scratch on annotated target domain data (referred as \textit{Seg-CT});
3) fine-tuning the source domain segmenter with annotated target domain data, i.e., the supervised transfer learning (referred as \textit{Seg-CT-STL});
4) directly testing the source domain segmenter on target domain data (referred as \textit{Seg-CT-noDA});
5) our proposed unsupervised domain adaptation method (referred as \textit{Seg-CT-UDA}).
We also compared with a previous state-of-the-art heart segmentation method using ConvNets~\cite{payermulti}.
Last but not least, we conducted ablation studies to observe how the adaptation depth would affect the performance.

\subsection{Results of Unsupervised Domain Adaptation}
The results of different methods are listed in Table~1, which demonstrates that the proposed unsupervised domain adaptation method is effective by mapping the feature space of target CT domain to that of source MRI domain. Qualitative results of the segmentations for CT images are presented in Fig.~3.

We first evaluate the performance of the segmenter for \textit{Seg-MRI}, which is the source domain model and serves as the basis for subsequent domain adaptation procedures. Compared with the~\cite{payermulti}, our ConvNet segmenter reached promising performance with exceeding Dice on LV-blood and LV-myo, as well as comparable Dice on AA and LA-blood. With this standard segmenter network architecture, we conducted following experiments to validate the effectiveness of our unsupervised domain adaptation framework.

To experimentally explore the potential upper-bounds of the segmentation accuracy of the cardiac structures from CT data, we implemented two different settings, i.e., the \textit{Seg-CT} and \textit{Seg-CT-STL}. Generally, the segmenter fine-tuned from \textit{Seg-MRI} achieved higher Dice and lower ASD than the model trained from scratch, proving the effectiveness of supervised transfer learning for adapting an established network to a related target domain using additional annotations. Meanwhile, these results are comparable to~\cite{payermulti} on most of the four cardiac structures.

As for observing the severe domain shift problem inherent in cross-modality biomedical images, we directly applied the segmenter trained on MRI domain to the CT data without any domain adaptation procedure. Unsurprisingly, the network of \textit{Seg-MRI} completely failed on CT images, with average Dice of merely 14.3\% across the structures. As shown in Table~1, the \textit{Seg-CT-noDA} only got a Dice of 0.8\% for the LV-blood. The model did not even output any correct predictions for two of the four testing subjects on the structure of LV-blood (please refer to (e) in Fig.~3). This demonstrates that although the cardiac MRI and CT images share similar high-level representations and identical label space, the significant difference in their low-level characteristics makes it extremely difficult for MRI segmenter to extract effective features for CT.

With our unsupervised domain adaptation method, we find a great improvement of the segmentation performance on the target CT data compared with the \textit{Seg-CT-noDA}. More specifically, our \textit{Seg-CT-UDA (d=21)} model has increased the average Dice across four cardiac structures by 43.4\%. 
As presented in Fig.~3, the predicted segmentation masks from \textit{Seg-CT-UDA} can successfully localize the cardiac structures and further capture their anatomical shapes.
The performance on segmenting AA is even close to that of \textit{Seg-CT-STL}. This reflects that the distinct geometric pattern and the clear boundary of the AA have been successfully captured by the DCM. In turn, it supervises the DAM to generate similar activation patterns as the source feature space via adversarial learning.
Looking at the other three cardiac structures (i.e., LA-blood, LV-blood and LV-myo), the \textit{Seg-CT-UDA} performances are not as high as that of AA.
The reason is that these anatomical structures are more challenging, given that they come with either relatively irregular geometrics or limited intensity contrast with surrounding tissues.
The deficiency focused on the unclear boundaries between neighboring structures or noise predictions on relatively homogeneous tissues away from the ROI. This is responsible for the high ASDs of \textit{Seg-CT-UDA}, where boundaries are corrupted by noisy outputs. Nevertheless, by mapping the feature space of target domain to that of the source domain, we obtained greatly improved and promising segmentations against \textit{Seg-CT-noDA} with zero data annotation effort.

\subsection{Ablation Study on Adaptation Depth}
The adaptation depth $d$ is an important hyper-parameter in our framework, which determines how many layers to be replaced during the plug-and-play domain adaptation procedure.
Intuitively, a shallower DAM (i.e., smaller $d$) might be less capable of learning effective feature mapping function $\mathcal{M}$ across domains than a deeper DAM (i.e., larger $d$).
This is due to the insufficient capacity of parameters in shallow DAM, as well as the huge domain shift in feature distributions.
Conversely, with an increase in adaptation depth $d$, DAM becomes more powerful for feature mappings, but training a deeper DAM solely with adversarial gradients would be more challenging.
Towards this issue, we conducted ablation studies to demonstrate how the performance would be affected by $d$.

To validate above intuitions and search for an optimal $d$, we repeated the experiment with domain adaptation from MRI to CT by varying the $d=\{13, 21, 31\}$, while maintaining all the other settings the same.
Viewing the examples in Fig.~4, \textit{Seg-CT-UDA (d=21)} model obtained an approaching ground-truth segmentation mask for ascending aorta. The other two models also produced inspiring results capturing the geometry and boundary characteristics of \textit{AA}, validating the effectiveness of our unsupervised domain adaptation method.
From the Table~1, we can observe that DAM with a middle-level of adaptation depth (\textit{d=21}) achieved the highest Dice on three of the four cardiac structures, exceeding the other two models by a significant margin.
For the LA-blood, the three adaptation depths reached comparable segmentation Dice and ASD, and the \textit{d=31} model was the best.
Notably, the model of \textit{Seg-CT-UDA (d=31)} overall demonstrated superiority over the model with adaptation depth \textit{d=13}.
This shows that enabling more layers learnable helps to improve the domain adaptation performance on cross-modality segmentations.

\begin{figure}[t]
\centering
\label{fig:depth}
\includegraphics[width=0.49\textwidth]{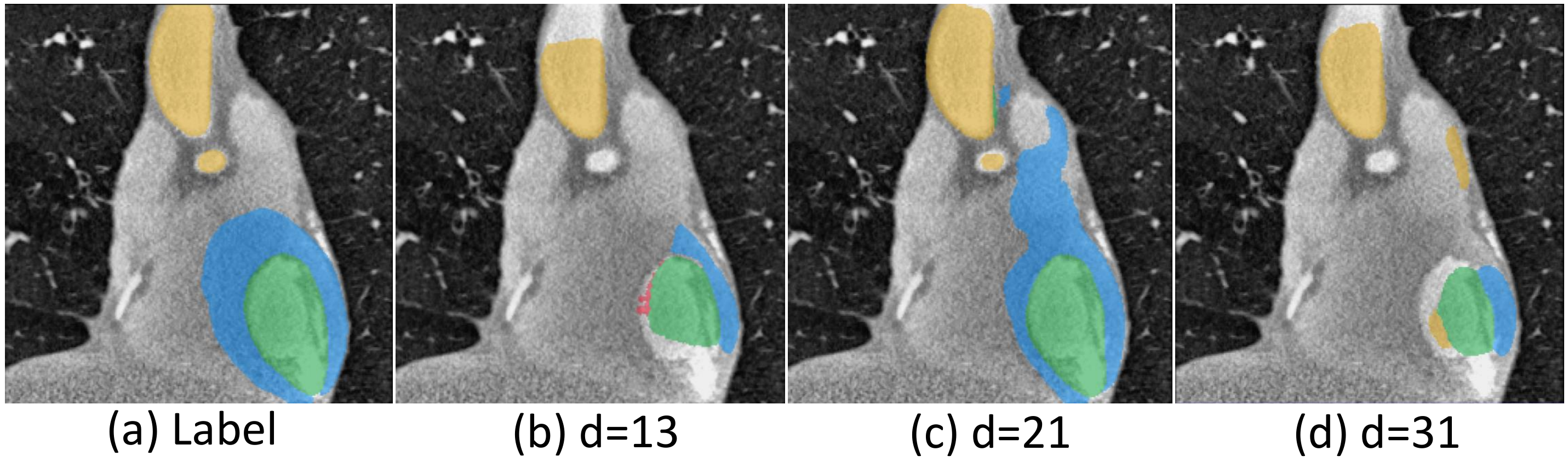}
\vspace{-7mm}
\caption{Comparison of results using \textit{Seg-CT-UDA} with different adaptation depth (colors are the same with Fig.~3).}
\vspace{-4mm}
\end{figure}

\vspace{-3mm}
\section{Conclusion}
This paper pioneers to propose an unsupervised domain adaptation framework for generalizing ConvNets across different modalities of biomedical images.
The flexible plug-and-play framework is obtained by optimizing a DAM and DCM via adversarial learning.
Extensive experiments with promising results on cardiac segmentations have validated the effectiveness of our approach.

\section*{Acknowledgments}
The work described in this paper was supported by the following grants from Hong Kong Research Grants Council under General Research Fund Scheme (Project no. 14202514 and 14203115).

\bibliographystyle{named}
\bibliography{refs}

\end{document}